\documentclass[runningheads]{llncs}

\usepackage{eccv}

\usepackage{eccvabbrv}

\usepackage{graphicx}
\usepackage{booktabs}

\usepackage[accsupp]{axessibility}  

\newcommand{\bs}[1]{\boldsymbol{#1}}

\newcommand{\hbs}[1]{\hat{\boldsymbol{#1}}}

\usepackage{makecell}
\usepackage{mathtools}
\DeclarePairedDelimiter{\norm}{\lVert}{\rVert}
\usepackage{wrapfig}

\usepackage{pifont}
\newcommand{\xmark}{\ding{55}}%

\usepackage{hyperref}

\usepackage{orcidlink}

\begin{document}
\emergencystretch 3em
\title{Fast Registration of Photorealistic Avatars for VR Facial Animation}


\author{Chaitanya Patel \inst{1} \and
Shaojie Bai \inst{2} \and
Te-Li Wang \inst{2} \and \\
Jason Saragih \inst{2} \and
Shih-En Wei \inst{2}}

\authorrunning{Chaitanya Patel~\etal}

\institute{Stanford University, USA\and
Meta Reality Labs, Pittsburgh, USA \\ \; \\
\url{https://chaitanya100100.github.io/FastRegistration/}
}

\maketitle

\begin{center}
    \centering
    \captionsetup{type=figure}
    \includegraphics[width=\textwidth]{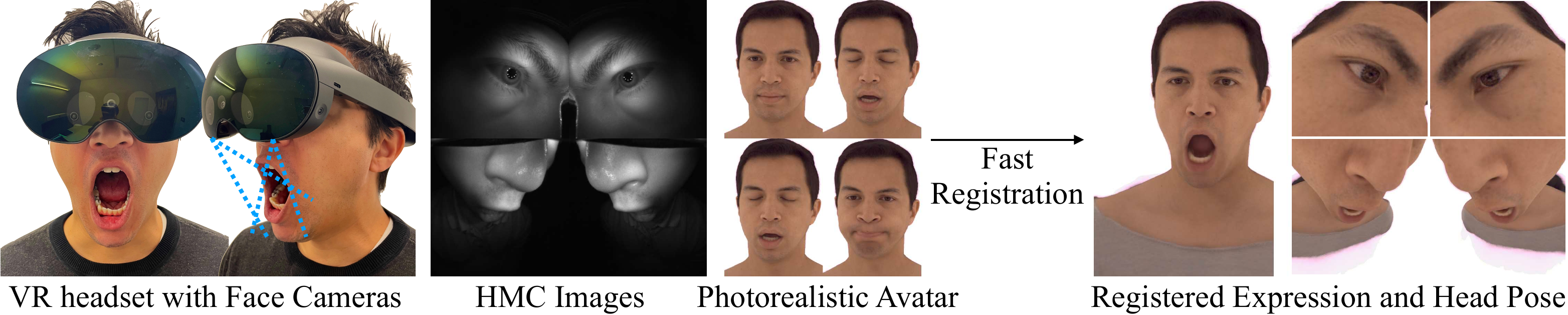}
    \captionof{figure}{On consumer VR headsets, \textbf{oblique mouth views and a large image domain gap hinder high quality face registration}. As shown, the subtle lip shapes and jaw movement are often hardly observed. Under this setting, our method is capable of efficiently and accurately registering facial expression and head pose of the photorealisitic avatars~\cite{chen2022instant} of unseen identities.}
\end{center}

\begin{abstract}

Virtual Reality (VR) bares promise of social interactions that can feel more immersive than other media. Key to this is the ability to accurately animate a personalized photorealistic avatar, and hence
the acquisition of the labels for headset-mounted camera (HMC) images need to be efficient and accurate, \textit{while} wearing a VR headset.
%
%
This is challenging due to oblique camera views and differences in image modality.
In this work, we first show that the domain gap between the avatar and HMC images is one of the primary sources of difficulty, where a transformer-based architecture achieves high accuracy on domain-consistent data, but degrades when the domain-gap is re-introduced.
Building on this finding, we propose a system split into two parts: an iterative refinement module that takes in-domain inputs, and a generic avatar-guided image-to-image domain transfer module conditioned on current estimates.
These two modules reinforce each other: domain transfer becomes easier when close-to-groundtruth examples are shown, and better domain-gap removal in turn improves the registration.
Our system obviates the need for costly offline optimization, and produces online registration of higher quality than direct regression method.
We validate the accuracy and efficiency of our approach through extensive experiments on a commodity headset, demonstrating significant improvements over these baselines.
To stimulate further research in this direction, we make our large-scale dataset and code publicly available.

\end{abstract}
\section{Introduction}
\label{sec:intro}

Photorealistic avatar creation has seen much progress in recent years. Driven by advances in neural representations and neural rendering~\cite{Lombardi2018deep,schwartz2020eyes,lombardi2019neural,Shysheya2019TexturedNA,applevisionpro}, highly accurate representations of individuals can now be generated even from limited captures such as phone scans~\cite{chen2022instant}
or monocular videos~\cite{giebenhain2023mononphm, applevisionpro}
while supporting real time rendering for interactive applications~\cite{wei2019VR, qian2023gaussianavatars}.
Photorealistic quality is achieved by learning a universal prior model~\cite{chen2022instant} on human appearance, which can be personalized to a novel user~\cite{chen2022instant,giebenhain2023mononphm}.
%
An emerging use case for such avatars is for enabling social interactions in Virtual Reality (VR). This application presents a particular problem where the user's face is typically occluded from the environment by the VR headset. As such, it relies on headset-mounted cameras (HMCs) to animate a user's avatar.
While accurate results have been demonstrated, they have been restricted to \textit{person-specific} cases, where correspondence pairs between the avatar and HMC images are obtained using additional elaborate capture rigs~\cite{wei2019VR}.
%
%
Highly accurate tracking in the more \textit{general} case remains an open problem, due to the need of specializing a generic encoder to users' personalized avatars, \textit{while} user is wearing a VR headset. Although fast adaptation methods are well studied~\cite{guo2020learning, chen2021highfidelity, browatzki2020}, the unsolved challenge here is to obtain high quality image-label pair, especially under oblique camera angles, time constraints, and the image domain difference between HMC images and avatar renderings.
%

In this work, we demonstrate 
that generic facial expression registration can be both accurate and efficient on unseen identities and challenging viewing angles.
%
For this, we first demonstrate that accurate results are possible when the modalities of the headset-mounted cameras (typically infrared) and the user's avatar match, using a novel transformer-based network that iteratively refines expression estimation and head pose, solely from image features. Our method assumes no requirement on avatar to provide landmarks, which are not reliable under oblique HMC views.  
Building on this finding, we propose to learn a cross-identity style transfer function from the camera's domain to that of the avatar. The core challenge here lies in the high fidelity requirement of the style transfer due to the challenging viewpoints of the face presented by headset mounted cameras; even a few pixels error can lead to significant effects in the estimated avatar's expression. 
To resolve this, a key design of our method is to leverage an iterative expression and head pose estimation, as well as a style transfer module, which reinforce each other. On one hand, given a higher-quality style transfer module, the iterative refinement process gets increasingly easier. 
On the other hand, when a refined expression and pose estimation is closer to groundtruth, the style transfer network can easily reason locally using the input HMC images (conditioned on multiple \emph{reference} avatar renderings) to remove the domain gap. 

To demonstrate the efficacy of our approach, we perform experiments on a dataset of 208 identities, each captured in a multiview capture system~\cite{Lombardi2018deep} as well as a modified QuestPro headset~\cite{questpro}, where the latter was used to provide ground truth correspondence between the driving cameras and the avatars. Compared to direct regression method, our iterative construction shows significantly improved robustness against novel appearance variations in unseen identities.
%
%

In summary, the contribution of this work include:
\begin{itemize}
    \item A demonstration that accurate and efficient generic face registration on a neural rendering face model is achievable under matching camera-avatar domains, without relying on 3D geometry.
    \item A generalizing style transfer network that precisely maintains facial expression on unseen identities, conditioned on photorealistic avatar renderings.
    \item Overall, a method to establish high-fidelity image-label pairs for unseen personalized avatars under time constraints and oblique viewing angles.
\end{itemize}
The remaining of the paper is structured as follows. In the next section, a literature review is presented. Then, in \S\ref{sec:method}, we outline our method for generic facial expression estimation. In \S\ref{sec:exp}, we demonstrate the efficacy of our approach with extensive experiments. We conclude in \S\ref{sec:conclusions} with a discussion of future work.

\section{Related Work}
\label{sec:related}

\subsection{VR Face Tracking}
While face tracking is a long studied problem, tracking faces of VR users from head mounted cameras (HMCs) poses an unique challenge. The difficulty mainly comes from restrictions in camera placement and occlusion caused by the headset. Sensor images only afford oblique and partially overlapping views of facial parts.
Previous work explored different ways to circumvent these difficulties. In~\cite{Hao2015}, a camera was attached on a protruding mount to acquire a frontal view of the lower face, but with a non-ergonomic hardware design. In~\cite{FaceVR2018}, the outside-in third-person view camera limits the range of a user's head pose. 
Both of these works rely on RGBD sensors to directly register the lower face with a geometry-only model. 
To reduce hardware requirements,~\cite{Olszewski2016} used a single RGB sensor for the lower face and performed direct regression of blendshape coefficients. The training dataset comprised of subjects performing a predefined set of expressions and sentences that had associated artist-generated blendshape coefficients. The inconsistencies between subject's performances with the blendshape-labeled animation limited animation fidelity.

A VR face tracking system on a consumer headset (Oculus Rift) with photoreaslitic avatars~\cite{Lombardi2018deep} was firstly presented in~\cite{wei2019VR}. They introduced two novelties: (1) \textit{The concept of a training- and tracking-headset}, where the former has a super-set of cameras of the latter. After training labels were obtained from the \emph{training headset}, the auxiliary views from better positioned cameras can be discarded, and a regression model taking only \emph{tracking headset}'s input was built. They also employed (2) \textit{analysis-by-synthesis with differentiable rendering and style transfer} to precisely register parameterized photorealistic face models to HMC images, bridging the RGB-to-IR domain gap. 
The approach was extended in~\cite{schwartz2020eyes} via jointly learning the style-transfer and registration together, instead of an independent CycleGAN-based module. 
Although highly accurate driving was achieved, both~\cite{wei2019VR} and~\cite{schwartz2020eyes} relied on person-specific models, the registration process required hours to days of training, and required the \emph{training headset} with auxiliary camera views to produce ground truth. As such, they cannot be used in a live setting where speed is required and only cameras on consumer headsets are available.
In this work, we demonstrate that a system trained on a pre-registered dataset of multiple identities can generalize well to unseen identities' HMC captures within seconds. These efficiently generated image-label pairs can later be used to adapt a generic realtime expression regressor and make the animation more precise.

\subsection{Image Style Transfer}
The goal of image style transfer is to render an image in a target style domain provided by conditioning information, while retaining semantic and structural content from an input's content. Convolutional neural features started to be utilized~\cite{gatys2016image} to encode content and style information.
Pix2pix~\cite{pix2pix2017} learns conditional GANs along with $L_1$ image loss to encourage high-frequency sharpness, with an assumption of availability of paired ground truth. 
To alleviate the difficulty of acquiring paired images, CycleGAN~\cite{CycleGAN2017} introduced the concept of cycle-consistency, but each model is only trained for a specific pair of domains, and suffers from semantic shifts between input and output.
StarGAN~\cite{choi2018stargan} extends the concept to a fixed set of predefined domains.
For more continuous control, many explored text conditioning~\cite{brooks2023instructpix2pix} or images conditioning~\cite{deng2021stytr2,wu2021styleformer,chen2021artistic,liu2021adaattn,an2021artflow}. These settings usually have information imbalance between input and output space, where optimal output might not be unique.
In this work, given a latent-space controlled face avatar~\cite{chen2022instant}, along with a ground-truth generation method~\cite{schwartz2020eyes}, our style transfer problem can simply be directly supervised, with conditioning images rendered from the avatar to address the imbalance information problem.

\subsection{Learning-based Iterative Face Registration}



A common approach for high-precision face tracking involves a cascade of regressors that use image features extracted from increasingly registered geometry. 
One of the first methods to use this approach used simple linear models raw image pixels~\cite{saragih2006ebm}, which was extended by using SIFT features~\cite{Xiong2013SupervisedDM}. 
Later methods used more powerful regressors, such as binary  trees~\cite{cao2014DDE,Kazemi_2014_CVPR} 
and incorporated the 3D shape representation into the formulation. Efficiency could be achieved by binary features and linear models~\cite{ren2014face3000}.

While these face tracking methods use current estimates of geometry to extract relevant features from images, similar cascade architectures have also been explored for general detection and registration. In those works, instead of \emph{extracting} features using current estimates of geometry, the input data is augmented with \emph{renderings} of the current estimate of geometry, which simplifies the backbone of the regressors in leveraging modern convolutional deep learning architectures. For example, Cascade Pose Regression~\cite{Dollar2010cascaded} draws 2D Gaussians centered at the current estimates of body keypoints, which are concatenated with the original input, acting as a kind of soft attention map. Similar design in~\cite{Bulat_2017_ICCV} was used for 3D heatmap prediction.
%
Xia et al.~\cite{xia2022sparse} applied vision transformer~\cite{Dosovitskiy2020AnII} to face alignment with landmark queries. 
In this work, we demonstrate a transformer-based network that doesn't require any guidance from landmark to predict precise corrections of head pose and expression from multiview images.
\section{Method}
\label{sec:method}

\begin{figure}[t]
\centering
\includegraphics[width=\textwidth]{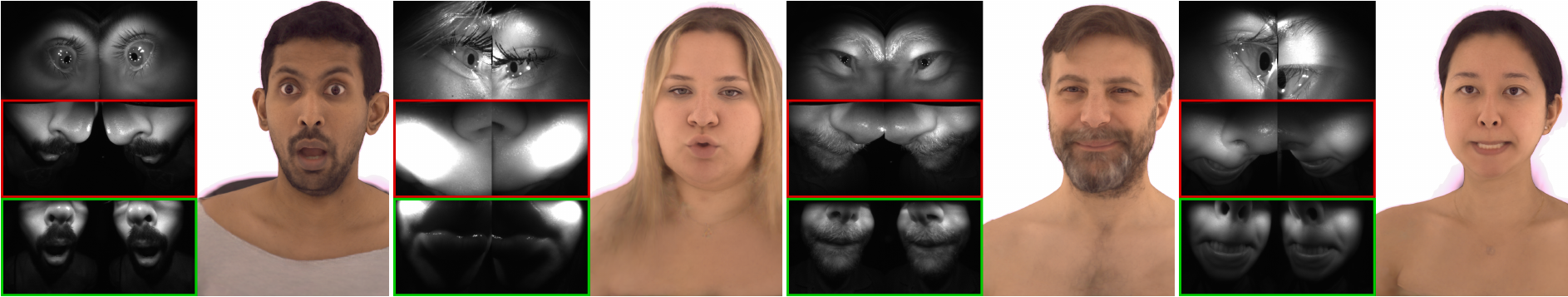}
\caption{Examples of HMC images and corresponding ground truth expression rendered on their avatars from the offline registration method~\cite{schwartz2020eyes}, which utilizes augmented cameras with better frontal views (highlighted in \textcolor{green}{green}). In this work, we aim to efficiently register faces using cameras on consumer headsets, which only have oblique views (highlighted in \textcolor{red}{red}). In such views, information about subtle expressions (e.g., lip movements) are often covered by very few pixels or even not visible. }
\label{fig:hmc_example}
\vspace{-4mm}
\end{figure}

We aim to register the avatar face model presented in~\cite{chen2022instant} to multi-view HMC images denoted
$\bs H = \{H_c\}_{c\in C}$, where each camera view $H_c \in \mathbb{R}^{h\times w}$ is a monochrome infrared (IR) image and $C$ is the set of available cameras on a consumer VR headset (in this work, we primarily focus on Meta's Quest Pro~\cite{questpro}, see the supplementary material). They comprise a patchwork of non-overlapping views between each side of the upper and lower face. Some examples are shown in Fig.~\ref{fig:hmc_example}. Due to challenging camera angles and headset donning variations, it is difficult for the subtle facial expressions to be accurately recognized by machine learning models (e.g., see Fig.~\ref{fig:sdm_results}).  

\paragraph{Setting.} We denote the avatar's decoder model from~\cite{chen2022instant} as $\mathcal{D}$. Following the same setting as in~\cite{chen2022instant}, given an input expression code $\bs z \in \mathbb{R}^{256}$, viewpoint $\bs v \in \mathbb{R}^6$, and identity information of the $i^\text{th}$ subject, $\bs{I}^i$, the decoder is able to render this subject's avatar from the designated viewpoint by $R = \mathcal{D}(\bs z, \bs{v} | \bs{I}^i) \in \mathbb{R}^{h \times w \times 3}$.
Specifically, when we use $\bs{v}=\bs{v}_c$; i.e., the viewpoint of a particular head-mounted camera (HMC), we'll obtain $R_c=\mathcal{D}(\bs z, \bs{v}_c | \bs{I}^i) \in \mathbb{R}^{h \times w \times 3}$, which has the same view as the corresponding $H_c \in \mathbb{R}^{h \times w}$, except the latter is monochromatic.
%
%
Following~\cite{chen2022instant}, the identity information $\bs{I}^i$ for a specific identity $i$ is provided as multi-scale untied bias maps to the decoder neural network. In this paper, we assume $\bs{I}^i$ is available for both training and testing identities, either from the lightstage or a phone scanning\footnote{In this work we differentiate between unseen identities for avatar generation vs. unseen identities for HMC driving. We always assume an avatar for a new identity is already available through prior works, and evaluate the performance of expression estimation methods on unseen HMC images of that identity.}; and that the calibrations of all head-mounted cameras are known. 
We utilize the method in~\cite{schwartz2020eyes} to establish groundtruth HMC image-to-($\bs z$,$\bs v$) correspondences, which relies on an identity-specific costly optimization process and an augmented additional camera set, $C'$, which provides enhanced visibility. The examples are highlighted in the green boxes in Fig.~\ref{fig:hmc_example}.
%
Our goal in this work is to estimate the same optimal $\bs z$ and $\bs v$ for new identities leveraging the avatar model (i.e., registration), while using only the original camera set $C$, highlighted in red boxes in Fig.~\ref{fig:hmc_example}.

\subsection{A Simplified Case: Matching Input Domain}
\label{subsec:contrived}
Accurate VR face registration entails exact alignment between $H_c$ and $R_c$ for each head-mounted camera $c$. However, a vital challenge here is their enormous domain gap: $\bs H=\{H_c\}_{c \in C}$ are monochrome infrared images with nearfield lighting and strong shadows, while $\bs R=\{R_c\}_{c \in C}$ are renderings of an avatar built from uniformly lit colored images in the visible spectrum.
\cite{wei2019VR, schwartz2020eyes} utilized a style transfer network to bridge this gap in a identity-specific setting.
To simplify the problem in the generic, multi-identity case, we first ask the question: what performance is possible when there is no domain difference? 
To study this, we replace $\bs{H}$ with $\bs{R}_{gt} = \mathcal{D}(\bs z_{gt}, \bs v_{gt})$ obtained from the costly method in~\cite{schwartz2020eyes} with augmented cameras.
$\bs{R}_{gt}$ can be seen as a perfectly style transferred result from $\bs{H}$ to the 3D avatar rendering space, that exactly retains expression.
To predict $(\bs z_{gt}, \bs v_{gt})$ from $\bs{R}_{gt}$, a na\"ive way is to build a regression CNN which can be made extremely efficient such as MobileNetV3~\cite{howard2019searching}.
Alternatively, given $\mathcal{D}$ is differentiable and the inputs are in the same domain, another straightforward approach is to optimize $(\bs z, \bs v)$ to fit to $\bs R_{gt}$ using pixel-wise image losses. As we show in Table~\ref{tab:contrived}, the regression model is extremely lightweight but fails to generalize well; whereas this offline method (unsurprisingly) generates low error, at the cost of extremely long time to converge.
%
%
Note that despite the simplification we make on the input domain difference (i.e., assuming access to $\bs{R}_{gt}$ rather than $\bs{H}$), the registration is still challenging due to the inherent oblique viewing angles, headset donning variations and the need to generalize to unseen identities. 

\begin{table}[t]\centering \footnotesize
\setlength{\tabcolsep}{5.0pt}
\caption{Registration accuracy in a simplified setting. The errors are averaged across all frames in the test set. Augmented cameras means the use of camera set $C'$ (which has better lower-face visibility) instead of $C$. Frontal Image $L_1$ describes expression prediction error, while rotation and translation errors describe the headpose prediction error. All methods are compared against groundtruth generated by the offline method~\cite{schwartz2020eyes} trained \textit{with augmented cameras}. *Note that offline method below (colored in \textcolor{lightgray}{gray}) is computed without augmented cameras, and is impractical due to the long convergence time.}
\begin{tabular}{l|cccccc}
\toprule
& \thead{Aug.\\Cams} & Input & \thead{Frontal\\ Image $L_1$} & \thead{Rot. Err. \\ (deg.)} &  \thead{Trans. Err.\\(mm)} & Speed \\
\cmidrule{1-7}
\textcolor{lightgray}{Offline~\cite{schwartz2020eyes}}* & \textcolor{lightgray}{\xmark} & \textcolor{lightgray}{$\bs{R}_{gt}$} & \textcolor{lightgray}{$0.784$}  & \textcolor{lightgray}{$0.594$} & \textcolor{lightgray}{$0.257$} & \textcolor{lightgray}{$\sim$1 day} \\
Regression & \xmark & $\bs{R}_{gt}$ & $2.920$ & $3.150$ & $2.900$ & \textbf{7ms} \\
Regression & \checkmark & $\bs{R}_{gt}$ & $2.902$ & $3.031$ & $3.090$ & \textbf{7ms} \\
\textbf{Ours} $\mathcal{F}_0$ & \xmark & $\bs{R}_{gt}$ & $\bs{1.652}$ & $\bs{0.660}$ & $\bs{0.618}$ & 0.4sec \\
\textbf{Ours} $\mathcal{F}_0$ & \checkmark & $\bs{R}_{gt}$ & $\bs{1.462}$ & $\bs{0.636}$ & $\bs{0.598}$ & 0.4sec \\
\cmidrule{1-7}
\textbf{Ours} $\mathcal{F}_0$ & \xmark & $\bs H$ & $2.851$ & $1.249$ & $1.068$  & 0.4sec \\
\bottomrule
\end{tabular}
\label{tab:contrived}
\end{table}


In contrast, we argue that a carefully designed function that leverages avatar model (i.e., $\mathcal{D}$) information, which we denote as $\mathcal{F}_{0}(\cdot | \mathcal{D})$, achieves a good balance: (1) it is feed-forward (no optimization needed for unseen identities) so its speed can afford online usage; (2) it utilizes the renderings of $\mathcal{D}$ as a feedback to compare with input $H_c$ and minimize misalignment.
Before we describe $\mathcal{F}_{0}$ in \S~\ref{subsec:sdm},
we report the results of aforementioned methods under this simplified setting in Table~\ref{tab:contrived}. 

Specifically, we show that $\mathcal{F}_{0}$ can achieve performance approaching that of offline registration~\cite{schwartz2020eyes}. In contrast, na\"ive direct regressions perform substantially worse, even with the augmented set of cameras.
This highlights the importance of conditioning face registration learning with information about the target identity's avatar (in our case, $\mathcal{D}$). But importantly, when reverting back to the real problem, by replacing $\bs{R}_{gt}$ with $\bs{H}$, the performance of $\mathcal{F}_{0}$ also degrades significantly. 
This observation demonstrates the challenge posed by input domain gap difference, and motivates us to decouple style transfer problem from registration, as we describe next.

%
%


\subsection{Overall Design}

%
In light of the observation in \S\ref{subsec:contrived}, we propose to decouple the problem into the learning of two modules: an iterative refinement module, $\mathcal{F}$, and a style transfer module, $\mathcal{S}$.
The goal of $\mathcal{F}$ is to produce an iterative update to the estimate expression $\bs z$ and headpose $\bs v$ of a given frame.
However, as Table~\ref{tab:contrived} shows, conditioning on avatar model $\mathcal{D}$ alone is not sufficient; good performance of such $\mathcal{F}$ relies critically on closing the gap between $\bs{H}$ and $\bs{R}_{gt}$.
Therefore, module $\mathcal F$ shall rely on style transfer module $\mathcal S$ for closing this monochromatic domain gap. Specifically, in addition to raw HMC images $\bs H$, we also feed a style transferred version of them (denoted $\hat{\bs{R}}$), produced by $\mathcal{S}$, as input to $\mathcal F$. Intuitively, $\hat{\bs{R}}$ should then resemble avatar model $\mathcal{D}$'s renderings with the same facial expression as in $\bs H$. (And as Table~\ref{tab:contrived} shows, if $\hat{\bs{R}} \approx \bs{R}_{gt}$, one can obtain really good registration.)
Differing from the common style transfer setting, here the conditioning information that provides ``style'' to $\mathcal{S}$ is the entire personalized model $\mathcal{D}(\cdot|\bs{I}^i)$ itself. 
As such, we have the options of providing various conditioning images to $\mathcal{S}$ by choosing expression and viewpoints to render. Throughout experiments, we find that selecting frames with values closer to $(\bs z_{gt}, \bs v_{gt})$ improves the quality of $\mathcal{S}$'s style transfer output.

\begin{figure}[t]
\centering
\includegraphics[width=0.6\textwidth]{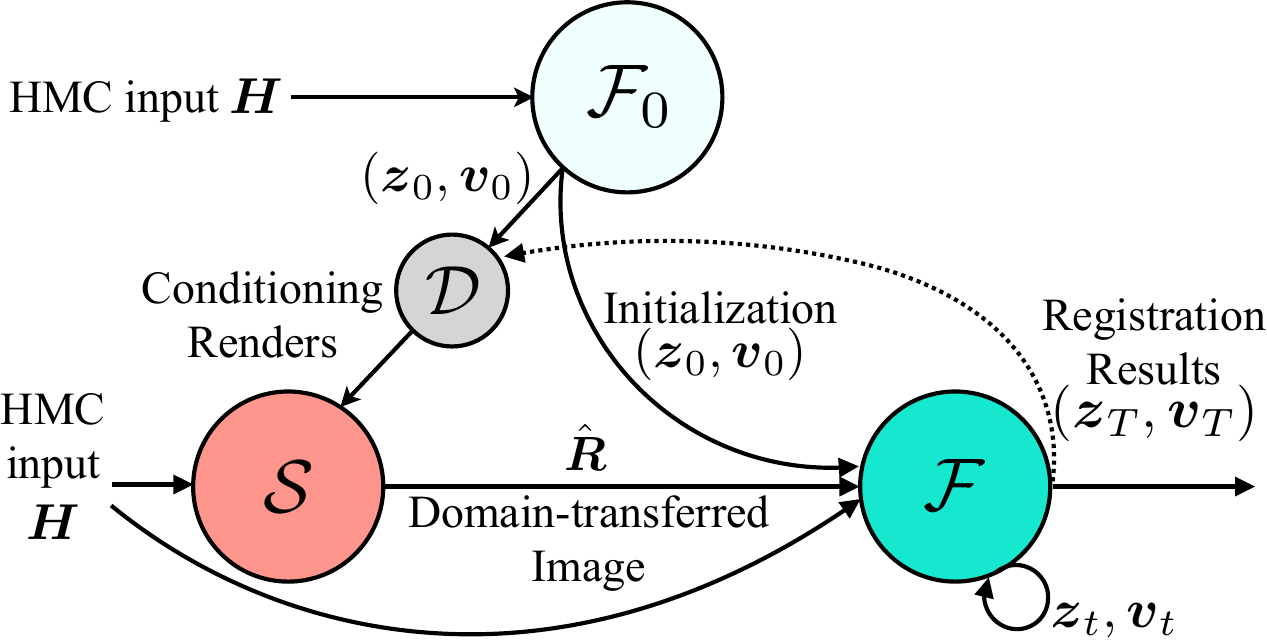}
\caption{Overview of the method. We decouple the problem into an avatar-conditioned image-to-image style transfer module $\mathcal{S}$ and a iterative refinement module $\mathcal{F}$. Module $\mathcal{F}_0$ initializes both modules by directly esimating on HMC input $\bs H$.}
\label{fig:overview}
\end{figure}

Therefore, a desirable mutual reinforcement is formed: the better $\mathcal{S}$ performs, the lower the errors of $\mathcal{F}$ are on face registration; in turn, the better $\mathcal{F}$ performs, the closer rendered conditioning images will be to the ground truth, simplifying the problem for  $\mathcal{S}$.
An initialization $(\bs z_0, \bs v_0) = \mathcal{F}_0(\bs H)$ for this reinforcement process can be provided by any model that directly works on monochromatic inputs $\bs H$.
Fig.~\ref{fig:overview} illustrates the overall design of our system.
In what follows, we will describe the design of each module.

\subsection{Transformer-based Iterative Refinement Network}
\label{subsec:sdm}

\begin{figure}[t]
\centering
\includegraphics[width=\textwidth]{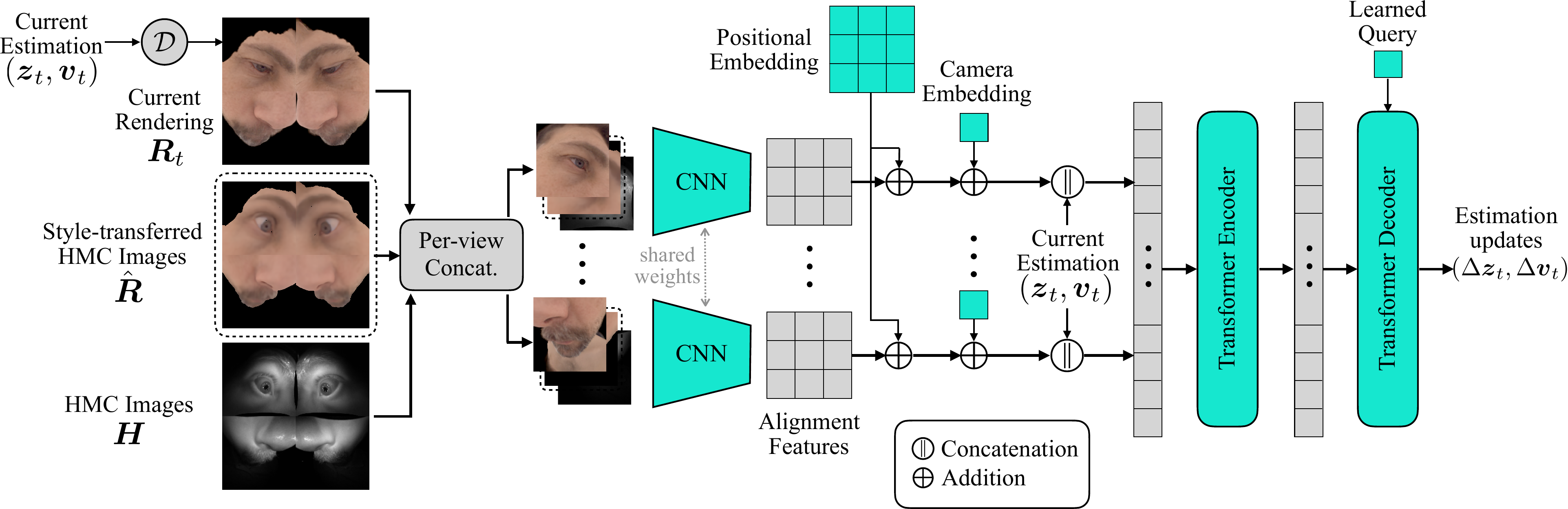}
\caption{Architecture of iterative refinement module $\mathcal{F}$}
\label{fig:sdm_arch}
\end{figure}

The role of the iterative refinement module, $\mathcal{F}$, is to predict the updated parameters $(\bs z_{t+1}, \bs v_{t+1})$ from input and current rendering:
\begin{align}
\label{eq:update}
    [\bs{z}_{t+1}, \bs{v}_{t+1}] &= \mathcal{F}\left(\bs{H}, \hbs{R}, \bs{R}_t \right), \ \ \bs{R}_t=\mathcal{D}(\bs{z}_{t}, \bs{v}_{t})
\end{align}
where $t \in [1, T]$ is number of steps and $\hbs{R} = \mathcal{S}(\bs{H})$ is the style-transferred result (see Fig.~\ref{fig:sdm_arch}). 
%
$\mathcal{F}$ can reason about the misalignment between input $\bs H$ and current rendering $\mathcal{D}(\bs{z}_{t}, \bs{v}_{t})$, with the aid of $\mathcal{S}(\bs{H})$ to bridge the domain gap.
%

\begin{figure}[t]
\centering
\includegraphics[width=\textwidth]{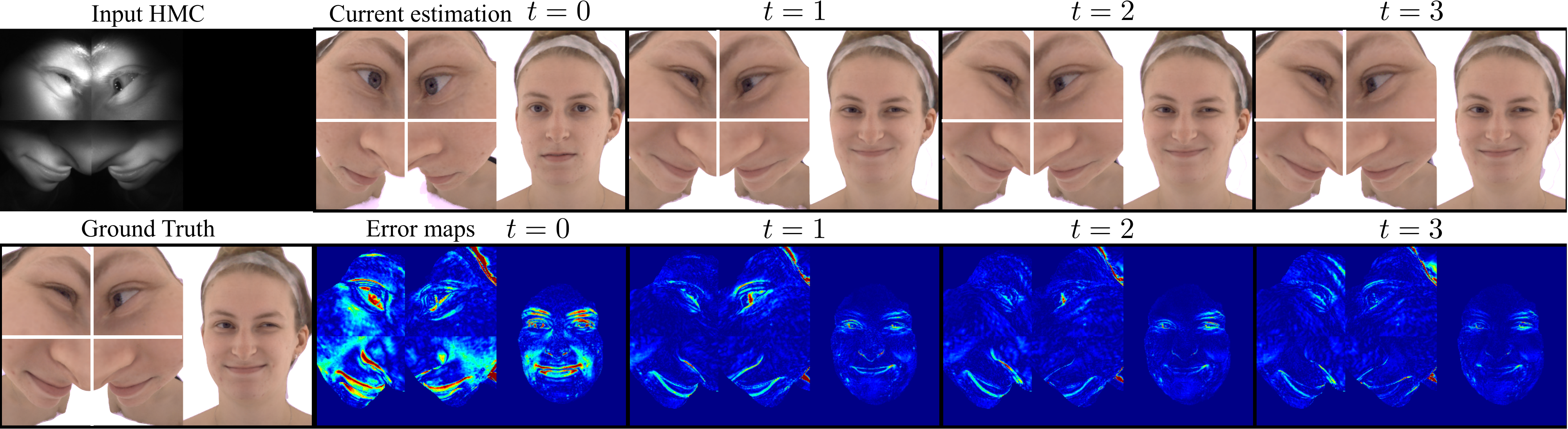}
\caption{Progression of iterative refinement in $\mathcal{F}$: we show intermediate results $\mathcal{D}(\bs{z}_{t}, \bs{v}_{t})$ and corresponding error maps for each step $t$.}
\label{fig:sdm_iters}
\end{figure}

In Fig.~\ref{fig:sdm_arch}, we show the hybrid-transformer~\cite{Dosovitskiy2020AnII} based architecture of $\mathcal{F}$.
For each view $c\in C$, a shared CNN encodes the alignment information between the current rendering $R_{t,c}$ and input images $H_c$ along with style-transferred images $\hat{R}_c$ into a feature grid. After adding learnable grid positional encoding and camera-view embedding, the grid features concatenated with the current estimate $(\bs z_t, \bs v_t)$ and are flattened into a sequence of tokens. These tokens are processed by a transformer module with a learnable decoder query to output $(\Delta \bs z_t, \Delta \bs v_t)$, which is added to $(\bs{z}_t, \bs{v}_t)$ to yield the new estimate for the next iteration.
We will show in \S\ref{subsec:ablation} that this hybrid-transformer structure is a crucial design choice for achieving generalization across identities. The transformer layers help to fuse feature pyramid from multiple camera views while avoiding model size explosion or information bottleneck.
%
Fig.~\ref{fig:sdm_iters} shows the progression of $\bs{R}_t$ over the steps.
This iterative refinement module is trained to minimize:
\begin{equation}
    \mathcal{L}_{\mathcal{F}} = \lambda_{\text{front}} \mathcal{L}_{\text{front}} + \lambda_{\text{hmc}} \mathcal{L}_{\text{hmc}},
\end{equation}
where 
\begin{align*}
\mathcal{L}_{\text{hmc}} &= \sum_{t=1}^{T} \sum_{c \in C} \norm{\mathcal{D}(\bs z_t, \bs v_{t,c}|\bs I^i) - \mathcal{D}(\bs z_{gt}, \bs v_{gt,c}|\bs I^i)}_1 \\
\mathcal{L}_{\text{front}} &= \sum_{t=1}^{T} \norm{\mathcal{D}(\bs z_t, \bs v_{\text{front}}|\bs I^i) - \mathcal{D}(\bs z_{gt}, \bs v_{\text{front}}|\bs I^i)}_1
\end{align*}
Here, $\bs v_{\text{front}}$ is a predefined frontal view of the rendered avatar (see Fig.~\ref{fig:sdm_iters}). While $\mathcal{L}_\text{hmc}$ encourages alignment between the predicted and groundtruth images from HMC views, $\mathcal{L}_{\text{front}}$ promotes an even reconstruction over the entire face to combat effects of oblique viewing angles in the HMC images. 


While $\mathcal{F}_0$ could be any module that works on HMC images $\bs H$ for the purpose of providing $\{\bs{z}_0, \bs{v}_0\}$, for consistency, we simply set $\mathcal{F}_0$ to also be iterative refining, where the internal module is the same as $\mathcal{F}$, except without $\hbs R$ as input.

\subsection{Avatar-conditioned Image-to-image Style Transfer}

The goal of the style transfer module, $\mathcal{S}$, is to directly transform raw IR input images $\bs H$ into $\hbs R$ that resembles the avatar rendering $\bs{R}_{gt}$ of that original expression. Our setting differs from the methods in the literature in that our style-transferred images need to recover identity-specific details including skin-tone, freckles, etc., that are largely missing in the IR domain; meanwhile, the illumination differences and oblique view angle across identities imply any minor changes in the inputs could map to a bigger change in the expression. These issues make the style transfer problem ill-posed without highly detailed conditioning.

\begin{figure}[t]
\centering
\includegraphics[width=0.8\textwidth]{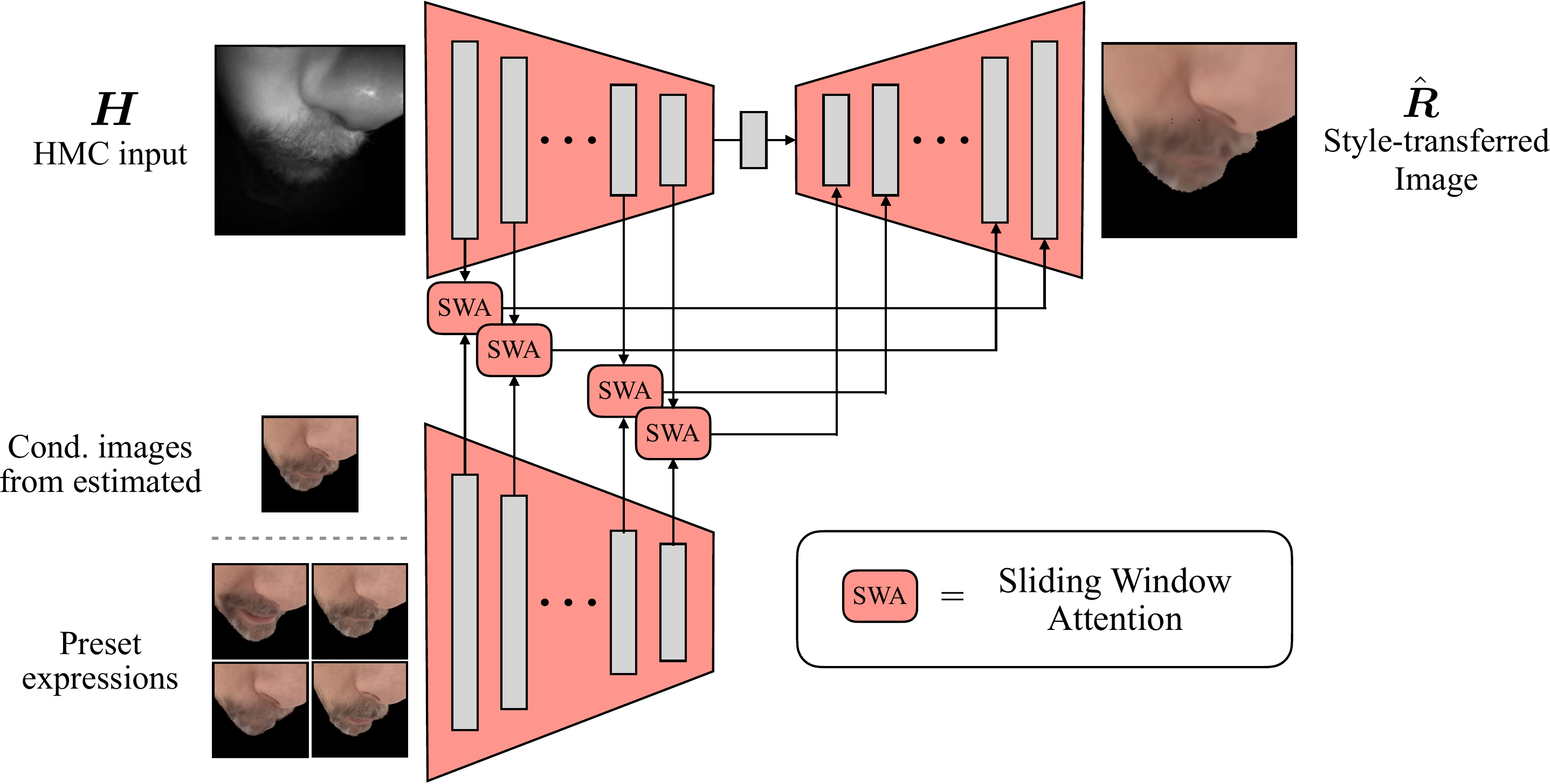}
\caption{Architecture of style transfer module $\mathcal{S}$}
\vspace{-4mm}
\label{fig:st_arch}
\vspace{-2mm}
\end{figure}

To this end, we design a novel style transfer architecture that utilizes the prior registration estimation given by $\mathcal{F}_0$. Specifically, we can utilize $\mathcal{F}_0$ that was trained directly on monochrome images $\bs{H}$, to obtain an estimate of $(\bs z_0, \bs v_0)$ for the current frame. Additionally, we choose $M$ \textit{reference conditioning expressions}: $(\bs z_{k_1},...,\bs{z}_{k_M})$ to cover a range of reference expressions; e.g., mouth open, squinting eyes, closed eyes, etc., which we find to significantly help mitigate ambiguities in style-transferring extreme expressions (we show examples of these conditioning reference expressions in the supplementary material). Formally, given the current frame HMC image $\bs{H}$, we compute
\begin{align}
\hbs R = \mathcal{S}\left(\bs H, (\bs z_0, \bs{z}_{k_1},...,\bs{z}_{k_M}), \bs{v}_0 \right).
\end{align}
%
%
With a better estimation of $(\bs z_0, \bs v_0)$ provided by $\mathcal{F}_0$, these conditioning images become closer to ground truth, thereby simplifying the style transfer learning task of $\mathcal{S}$.

Fig.~\ref{fig:st_arch} shows the UNet-based architecture of $\mathcal{S}$. Given an estimate of $(\bs z_0, \bs v_0)$, conditioning images are generated from the same estimate and $M$ other key expressions, concatenated channel-wise and encoded by a U-Net encoder. Input HMC image is encoded by a separate U-Net encoder. Sliding window based attention~\cite{sasa_parmar2019stand} modules are used to fuse input features and conditioning features to compensate for the misalignment between them. These fused features are provided as the skip connection in the U-Net decoder to output style-transferred image $\hbs R$.
This style transfer module is trained with a simple image $L_1$ loss:
\begin{equation}
\mathcal{L}_{\mathcal{S}} = \norm{\hbs R - \bs R_{gt}}_1. \\
\end{equation}

\section{Experiments}
\label{sec:exp}

We perform experiments on a dataset of 208 identities (1$M$ frames in total), each captured in a lightstage~\cite{Lombardi2018deep} as well as a modified Quest-Pro headset~\cite{questpro} with augmented camera views. 
%
The avatars are generated for all identities with a unified latent expression space using the method from~\cite{chen2022instant}. We utilize the extensive offline registration pipeline in~\cite{schwartz2020eyes}(with augmented camera set $C'$) to generate high-quality labels. 
We held out 26 identities as validation set. 
%
We use $T=3$ refinement iterations during training and $M=4$ key expressions to provide conditioning images for style transfer, which is operating at $192 \times 192$ resolution. See the supplementary material for more details on model architecture and training.




\subsection{Comparison with Baselines}

\begin{table}[t]\centering \footnotesize
\setlength{\tabcolsep}{5.0pt}
\caption{Comparison of our approach (with style transfer \emph{and} iterative refinement) with direct regression and offline methods. The errors are the averages of all frames in the test set. Augmented view means the use of camera set $C'$ instead of $C$. All methods are comparing against groundtruth generated by the offline method~\cite{schwartz2020eyes} trained \textit{with augmented cameras}. *Note that offline methods below (colored in \textcolor{lightgray}{gray}) are computed without augmented cameras, and are impractical due to the long convergence time. }
\begin{tabular}{l|cccccc}
\toprule
& \thead{Aug.\\cams} & Input & \thead{Frontal\\Image $L_1$} & \thead{Rot. Err.\\(deg.)} & \thead{Trans. Err.\\(mm)} & Speed \\
\cmidrule{1-7}
\textcolor{lightgray}{Offline~\cite{schwartz2020eyes}}* & \textcolor{lightgray}{\xmark} & \textcolor{lightgray}{$\bs H$} & \textcolor{lightgray}{$1.713$} & \textcolor{lightgray}{$2.400$} & \textcolor{lightgray}{$2.512$} & \textcolor{lightgray}{$\sim$1 day} \\
\textcolor{lightgray}{Offline~\cite{schwartz2020eyes}}* & \textcolor{lightgray}{\xmark} & \textcolor{lightgray}{$\bs{R}_{gt}$} & \textcolor{lightgray}{$0.784$}  & \textcolor{lightgray}{$0.594$} & \textcolor{lightgray}{$0.257$} & \textcolor{lightgray}{$\sim$1 day} \\
Regression & \xmark & $\bs H$ & $2.956$ & $2.850$ & $2.802$ & \textbf{7ms} \\
Regression & \xmark & $\bs{R}_{gt}$ & $2.920$ & $3.150$ & $2.900$ & \textbf{7ms} \\
Regression & \checkmark & $\bs H$ & $2.967$ & $2.806$ & $2.953$ & \textbf{7ms} \\
Regression & \checkmark & $\bs{R}_{gt}$ & $2.902$ & $3.031$ & $3.090$ & \textbf{7ms} \\
\cmidrule{1-7}
\textbf{Ours} ($\mathcal{F}$+$\mathcal{S})$ & \xmark & $\bs H$ & $\bs{2.655}$ & $\bs{0.947}$ & $\bs{0.886}$ & 0.4s \\
\textbf{Ours} ($\mathcal{F}$+$\mathcal{S})$ & \checkmark & $\bs H$ & $\bs{2.399}$ & $\bs{0.917}$ & $\bs{0.845}$ & 0.4s \\
\bottomrule
\end{tabular}
\label{tab:results}
\end{table}

\begin{figure}[t]
\centering
\includegraphics[width=\textwidth]{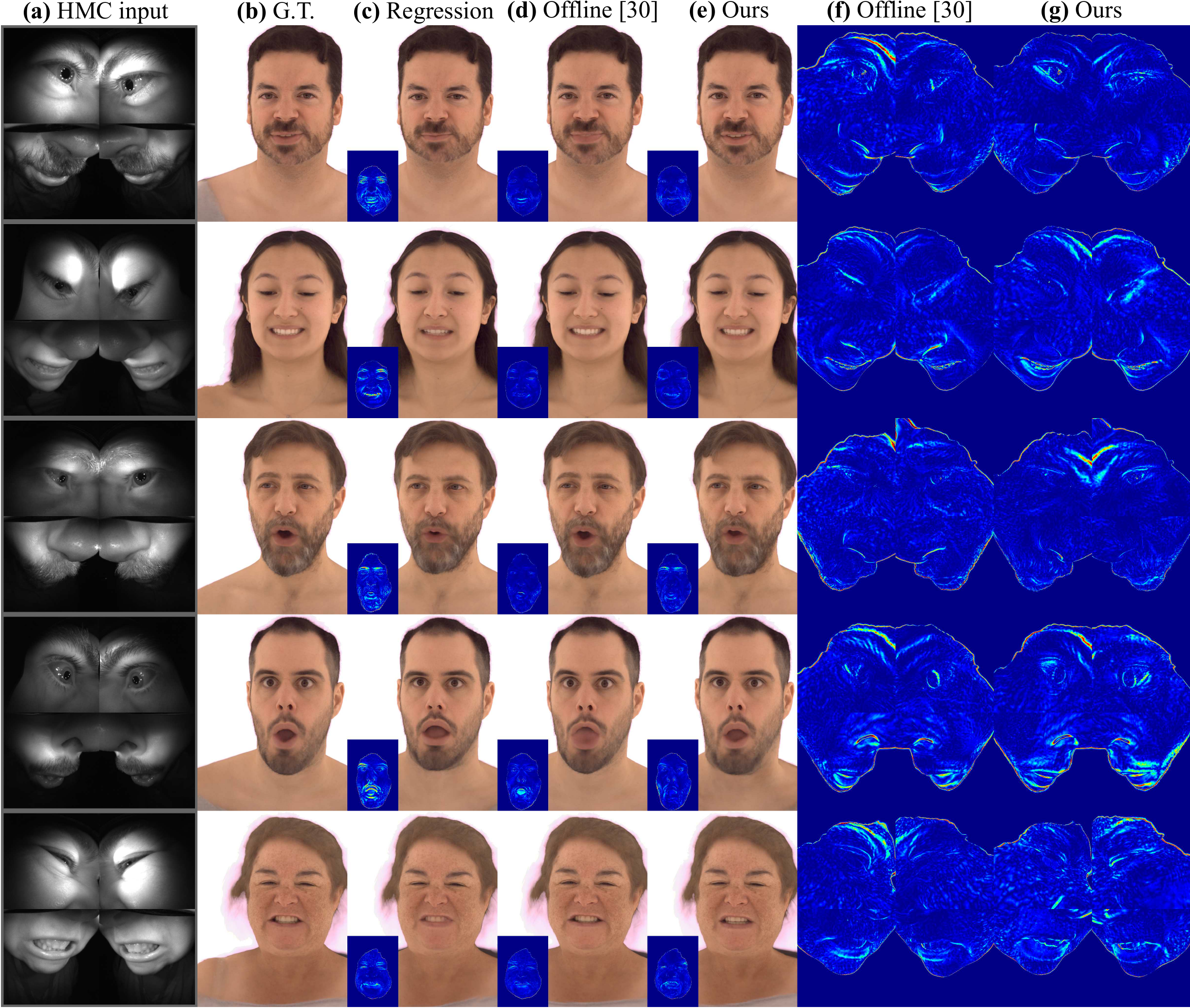}
\caption{\textbf{Qualitative Results:} we compare different methods by evaluating \textbf{(b,c,d,e)} frontal rendering (with error maps), and \textbf{(f,g)} error maps in HMC viewpoints. More examples are provided in the supplementary material.
}
\vspace{-4mm}
\label{fig:sdm_results}
\end{figure}

As discussed, there are two obvious types of methods to compare for general face registration: (1) the same \textbf{offline registration} method in~\cite{schwartz2020eyes}, but only using the camera set $C$, performed individually on each validation identity's headset data.
Since the training here is only across frames from that identity, it has the advantage to overfit on the same identity. However, it also limits the amount of prior knowledge it can leverage from other identities' images.
Its performance anchors the challenge from camera angles, if computing time is not limited.
%
(2) \textbf{Direct regression}: using the same set of ground truth labels, we train a MobileNetV3~\cite{howard2019searching} to directly regress HMC images to expression codes $\bs z$. This method represents an online model for a realtime system where iterative feedback is not possible because the use of $\mathcal{D}$ is prohibited.

Table~\ref{tab:results} summarizes the comparison. The offline method achieves good average frontal image loss. Albeit its high precision, it has common failure modes in lower jaw and inner mouth, where the observation is poor, as shown in Fig.~\ref{fig:sdm_results}. In comparison, our method could leverage the learning from cross-identity dataset, producing a more uniformly distributed error.
The offline method also suffers from worse head pose estimation because its co-optimized style transfer could compensate small errors in oblique viewing angle.
Our method is much faster due to its feed-forward design, enabling online generation of accurate labels.

On the other hand, the direct regression method generalizes poorly to novel identities, leading to worse performance on average.
It also yields inferior results in estimating head poses. 
The head pose is defined as the relative 3D transformation from a reference camera to the avatar center which is not consistent across identities and not observable from HMC images.
Since the regression baseline is not conditioned on avatar, there is no information to predict head poses accurately.
We also provide relaxed conditions (e.g. $\bs{R}_{gt}$ as input, or using augmented cameras), and interestingly it fails to improve, while our method can leverage these conditions significantly.
Our method's high accuracy, especially in the lip region as depicted in the supplementary video, captures nuanced facial expressions more effectively. These high-quality, quickly generated labels can be employed to adapt realtime regressors, thereby enhancing the immersive experience in virtual reality.

\subsection{Ablation Studies}
\label{subsec:ablation}

\begin{table}[t]\centering \footnotesize
\parbox{.47\linewidth}{
\centering
\caption{\textbf{Ablation on the design of $\mathcal{F}$}. All methods use $\bs R_{gt}$ as inputs and without augmented cameras. }
\begin{tabular}{l|ccc}
\toprule
&  \thead{Frontal\\Image\\$L_1$} & \thead{Rot.\\Err.\\(deg.)} &  \thead{Trans.\\Err.\\(mm)} \\
\cmidrule{1-4}
Ours $\mathcal{F}_0$ & $1.652$ & $0.660$ & $0.618$  \\
w/o transformer & 2.533 & 2.335 & 2.023  \\
w/o grid features & 2.786 & 2.818 & 3.081  \\
\makecell{w/o transformer \& \\w/o grid features} & 3.645 & 5.090 & 5.839  \\
\bottomrule
\end{tabular}
\label{tab:sdm_ablation}
}
\hfill
\parbox{.48 \linewidth}{
\centering
\caption{\textbf{Ablation on the design of $\mathcal{S}$}.}
\begin{tabular}{l|c}
\toprule
& \thead{Image $L_1$ Error} \\
\cmidrule{1-2}
Ours $\mathcal{S}$ & $2.55$ \\
w/o SWA & $2.82$ \\
\makecell{w/o key cond.\\expressions} & $2.75$ \\
w/o $\mathcal{F}_0$ & $2.99$  \\
\bottomrule
\end{tabular}
\label{tab:st_ablation}
}
\end{table}


\paragraph{Iterative Refinement Module $\mathcal{F}$.} 


Key to our design of $\mathcal{F}$ is the application of transformer on the grid of features from all camera views.
We validate this design by comparing the performance of $\mathcal{F}_0(\bs R_{gt})$ against the following settings:
\begin{itemize}
    \item \textbf{w/o transformer}, where we replace the transformer with an MLP. This approach is akin to our direct regression baseline but incorporates iterative feedback. Here, the grid features from all four camera views are simply concatenated and processed by an MLP. This trivial concatenation results in a 2x increase in the number of trainable parameters and subpar generalization.
    \item \textbf{w/o grid features}, where we average pool grid features to get a single feature for each camera view and use the same transformer design to process $|C|$ tokens.
    \item \textbf{w/o transformer \& w/o grid features}, where we use an MLP to process the concatenation of pooled features from all camera views.
\end{itemize}
Results are shown in Table~\ref{tab:sdm_ablation}. We can see that processing grid features using transformer results in better generalization while requiring fewer parameters compared to using an MLP with trivial concatenation.
Pooling grid features is also detrimental because it undermines minor variations in input pixels which are important in the oblique viewing angles of headset cameras. Transformer operating on grid tokens can effectively preserve fine-grained information and extract subtle expression details.

\paragraph{Style Transfer Module $\mathcal{S}$.}
We validate our design of $\mathcal{S}$ by comparing it with the following baselines:
\begin{itemize}
\item \textbf{w/o SWA}, where we simply concatenate the features of input branch with the features of conditioning branch at each layer.
\item \textbf{w/o key conditioning expressions}, where only the conditioning corresponding to the current estimate $(\bs z_0, \bs v_0)$ is used.
\item \textbf{w/o $\mathcal{F}_0$}, where conditioning is comprised only of the four key expressions rendered using the average viewpoint per-camera, $\bs v_{\text{mean}}$. 
\end{itemize}
Table~\ref{tab:st_ablation} shows the average $L_1$ error between the foreground pixels of the groundtruth image and the predicted style transferred image.
The larger error of style-transfer without $\mathcal{F}_0$ validates our design that a better style transfer can be achieved by providing conditioning closer to the groundtruth $(\bs z_{gt}, \bs v_{gt})$.
When not incorporating SWA or key conditioning expressions, the model performs poorly when the estimates $\bs v_0$ and $\bs z_0$ are suboptimal respectively, resulting in higher error.

\begin{figure}[t]
\centering
\includegraphics[width=0.7\textwidth]{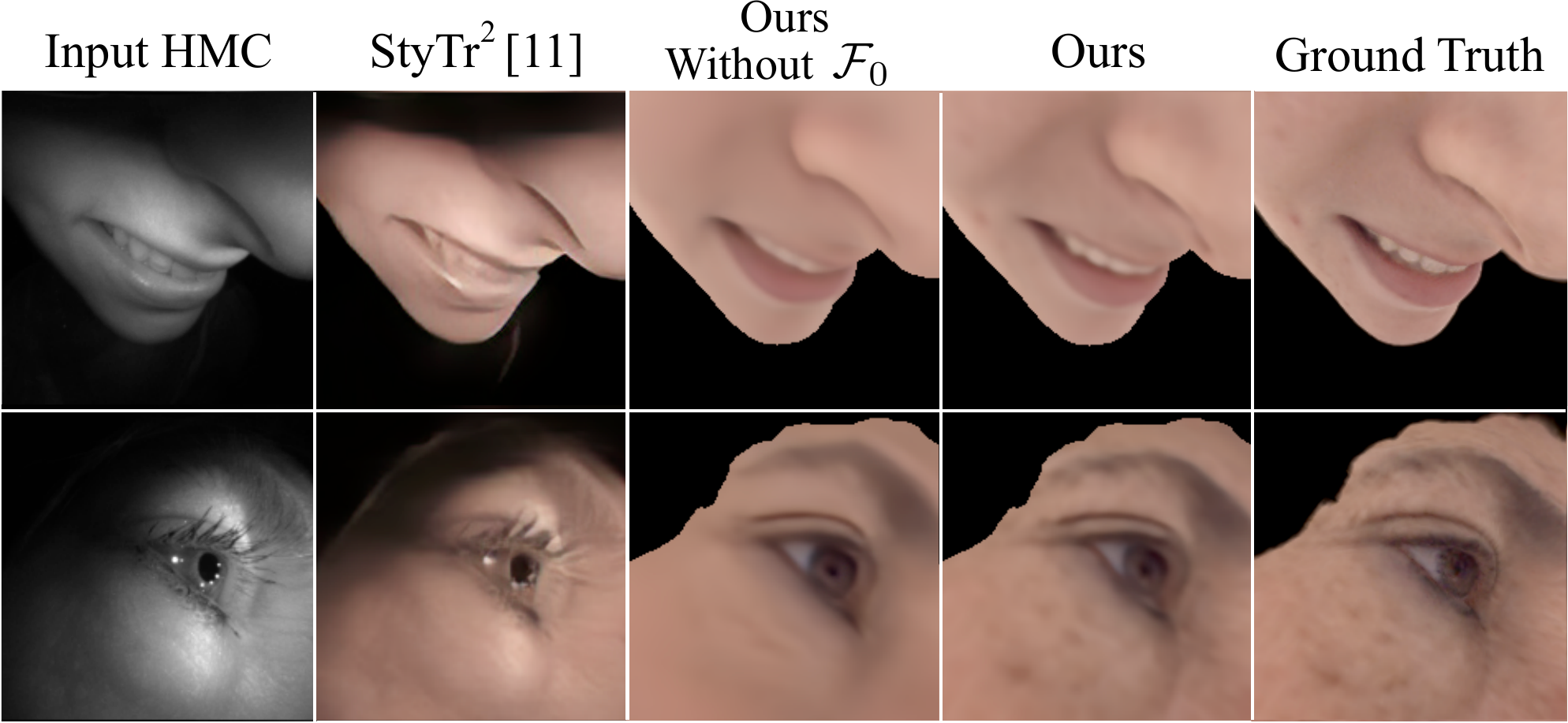}
\caption{\textbf{Ablation on style transfer results.} We compare our results with a generic style transfer method and baseline methods without the estimates provided by $\mathcal{F}_0$.} 
\label{fig:st_results}
\end{figure}

Fig.~\ref{fig:st_results} shows qualitative results of style transfer. 
Here, we also show the result of $\text{StyTr}^2$~\cite{deng2021stytr2} - one of the recent style transfer methods that leverages the power of vision transformers~\cite{Dosovitskiy2020AnII} with large datasets.
Despite using the groundtruth $\bs{R}_{gt}$ as the style image, it struggles to accurately fill in shadows and fine facial features that are not visible in the input HMC image.
Although `Without $\mathcal{F}_0$' produces better style transfer than $\text{StyTr}^2$~\cite{deng2021stytr2}, it smooths out high-frequency details including freckles, teeth, soft-tissue deformations near eyes and nose. These high-frequency details are crucial for animating subtle expressions. Our style transfer model $\mathcal{S}$ is able to retain such details by leveraging the estimate provided by $\mathcal{F}_0$. See the supplementary material for more results.


\section{Conclusion and Future Work}
\label{sec:conclusions}
In this paper, we present a generic and feed-forward method for efficient registration of photorealistic 3D avatars on monochromatic images with oblique viewing angles. 
We show that closing the domain gap between avatar's rendering and headset images is a key to achieve high registration quality. Motivated by this, we decompose the problem into two modules, style transfer and iterative refinement, and present a system where one reinforces the other. Extensive experiments on real capture data show that our system achieves superior registration quality than direct regression methods and can afford online usage. 
Our method provides a viable path for efficiently generating high quality label of neural rendering avatars on the fly, so that the downstream real-time model can adapt to achieve higher accuracy. This will enable the user to have photorealistic telepresence in VR without extensive data capture.
In the future, extensions of our method could be done for general registration of neural rendering models on out-of-domain multi-view images, such as (non-VR) face registration, body tracking, and 3D pose estimation.

%
%
\bibliographystyle{splncs04}
\bibliography{main}

\title{----- Supplementary Material ----- \\ Fast Registration of Photorealistic Avatars for VR Facial Animation} 
\titlerunning{Supplementary Material}

\author{Chaitanya Patel \inst{1} \and
Shaojie Bai \inst{2} \and
Te-Li Wang \inst{2} \and \\
Jason Saragih \inst{2} \and
Shih-En Wei \inst{2}}
\authorrunning{Chaitanya Patel~\etal}
\institute{Stanford University, USA\and
Meta Reality Labs, Pittsburgh, USA \\ \; \\
\url{https://chaitanya100100.github.io/FastRegistration/}
}
\maketitle
\setcounter{figure}{8}



\section{More Qualitative Results}

\begin{figure}[t]
\centering
\includegraphics[width=0.95\textwidth]{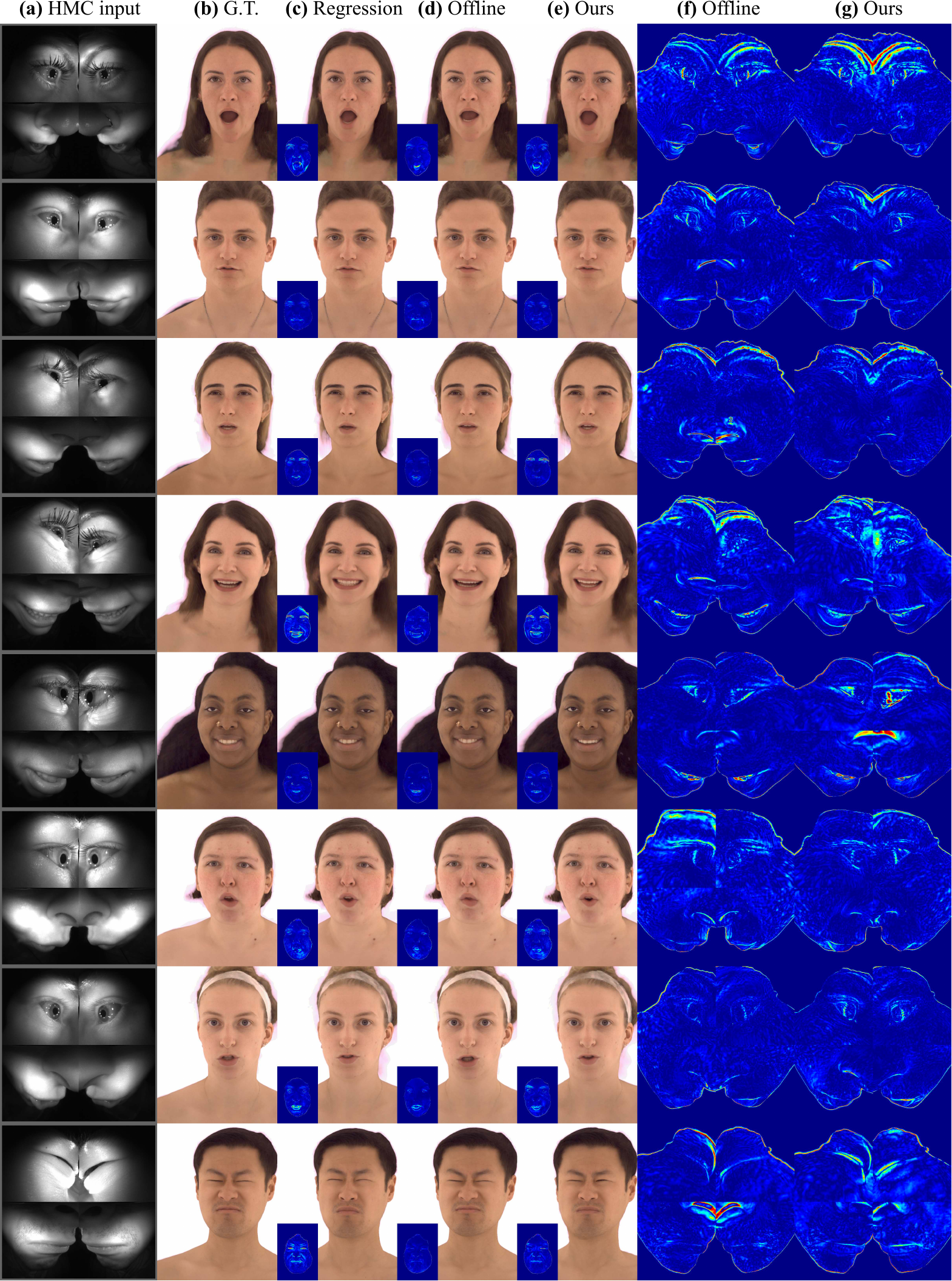}
\caption{\textbf{More Qualitative Results (1/2):} we compare different methods by evaluating \textbf{(b,c,d,e)} frontal rendering (with error maps), and \textbf{(f,g)} error maps in HMC viewpoints.}
\label{fig:more_qualitative_1}
\end{figure}

\begin{figure}[t]
\centering
\includegraphics[width=0.95\textwidth]{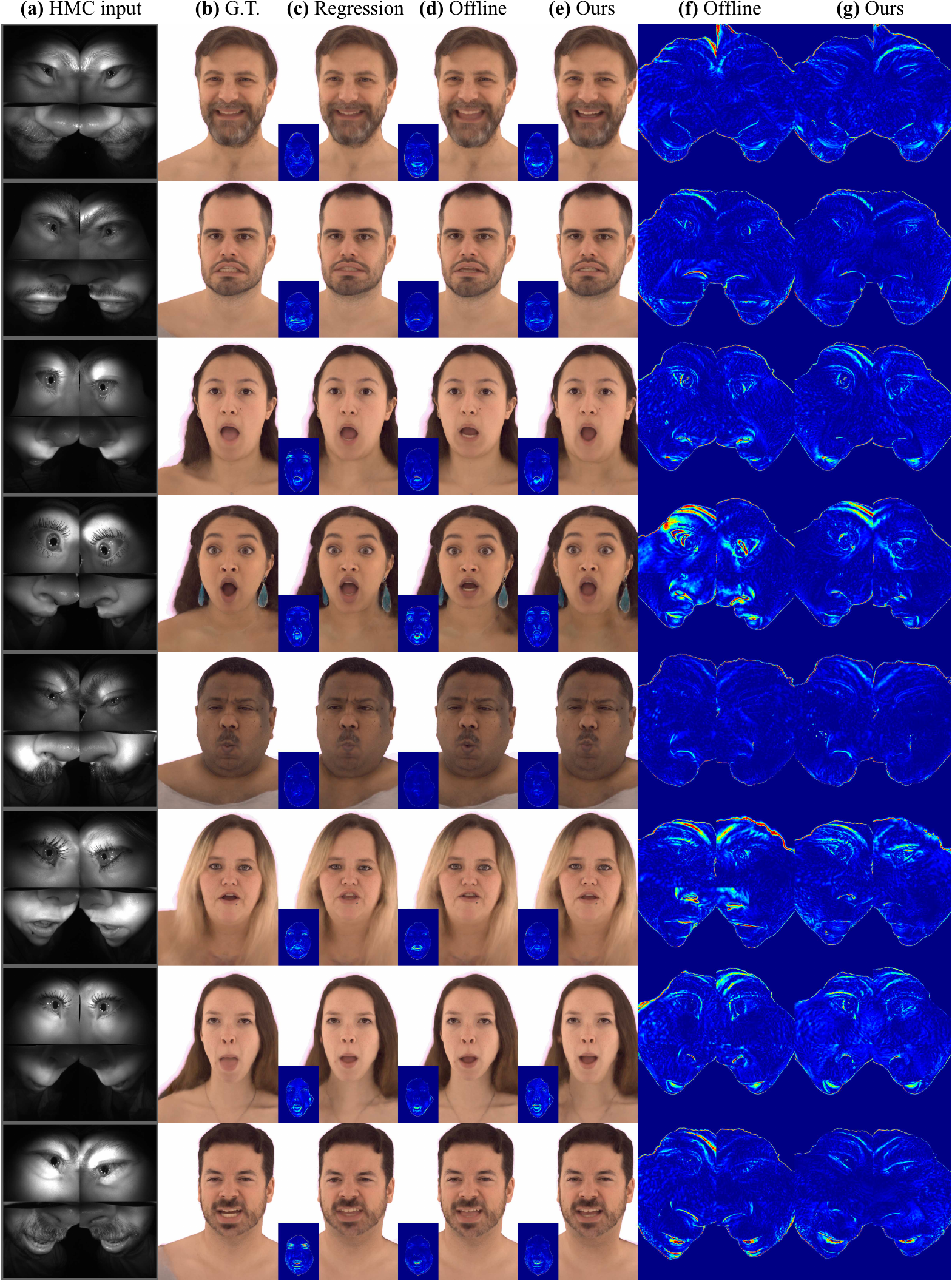}
\caption{\textbf{More Qualitative Results (2/2):} we compare different methods by evaluating \textbf{(b,c,d,e)} frontal rendering (with error maps), and \textbf{(f,g)} error maps in HMC viewpoints.}
\label{fig:more_qualitative_2}
\end{figure}

\begin{figure}[t]
\centering
\includegraphics[width=0.95\textwidth]{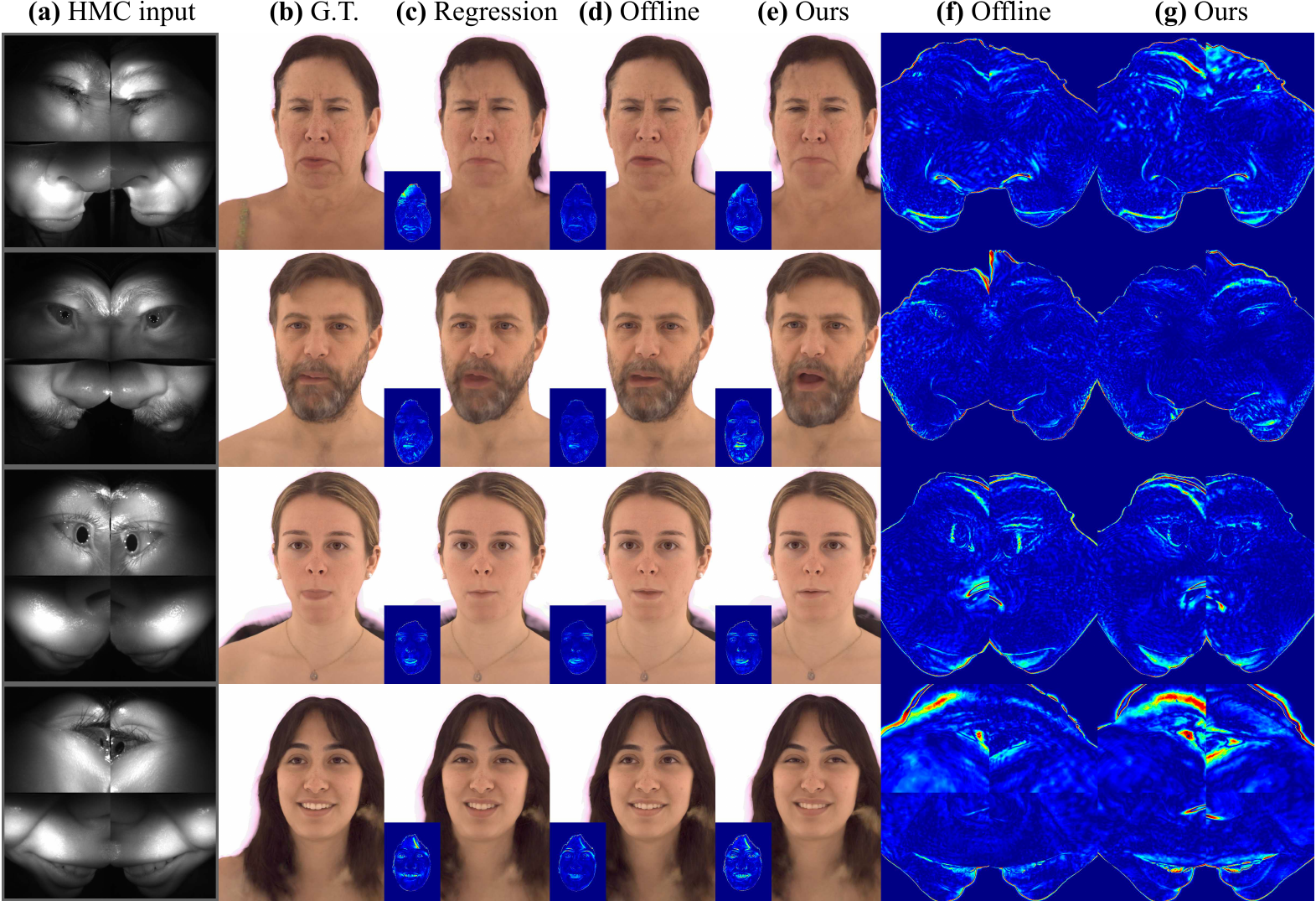}
\caption{\textbf{Failure cases of our methods:} we compare different methods by evaluating \textbf{(b,c,d,e)} frontal rendering (with error maps), and \textbf{(f,g)} error maps in HMC viewpoints.}
\label{fig:more_qualitative_fail}
\end{figure}

We show more qualitative results on test identities in Fig.~\ref{fig:more_qualitative_1} and Fig.~\ref{fig:more_qualitative_2} comparing against regression and offline method. More results can be found in the accompanying supplementary video.
Overall, the regression method has the larger error in terms of expression, often failing to capture subtle mouth shapes and the amount of teeth/tongue that is visible.
On the other hand, offline methods that slowly optimizes the expression code and head pose lead to lowest expression error overall. However, when key face areas are not well observed in the HMC images (e.g. row 1,3 in Fig.~\ref{fig:more_qualitative_1} and row 1,3,4,5,8 in Fig.~\ref{fig:more_qualitative_2}), our method often estimates better expressions. Our method is also superior in head pose estimation. For example, in row 4,6 of Fig.~\ref{fig:more_qualitative_2}, while our method has slightly high frontal error (expression), the offline method has higher head pose error, indicated by higher image error in the HMC perspective (column (f) and (g)). This is often caused by the style-transfer module compensating for registration error in its person-specific training regime~\cite{schwartz2020eyes} where the model can overfit more easily.  In contrast, our style transfer module is trained across a diverse set of identities, and does not overfit as easily, resulting in better retained facial structure, that in turn, leads to more accurate head pose. 
Fig.~\ref{fig:more_qualitative_fail} shows some failure cases of our method, which is usually caused by uncommon expressions, occluded mouth regions from HMC cameras, and extreme head poses. 
%


\section{Architecture Details}

\subsection{Iterative Refinement Module $\mathcal{F}$}
The iterative refinement module $\mathcal{F}$ has $\sim$28M trainable parameters. The CNN is based on ResNetV2-50~\cite{kolesnikov2020big} which takes as input images of size $128\times 128$ for each camera view and outputs $512\times 4\times 4$ grid features. After adding learnable patch embedding and view embedding, and concatenating the current estimate $(\bs z_t, \bs v_t)$, the sequence of $|C|\times 4\times 4$ feature tokens are processed by a ViT-based transformer module~\cite{Dosovitskiy2020AnII} that outputs the update $(\Delta\bs z_t, \Delta\bs v_t)$. The transformer module consists of 6 encoder layers and 4 decoder layers operating on 512-dim tokens. $\mathcal{F}_0$ follows the same architecture as $\mathcal{F}$ except without the style-transfer images $\hbs R$ as input.

\subsection{Style Transfer Module $\mathcal{S}$}

\begin{figure}[t]
\centering
\includegraphics[width=0.7\columnwidth]{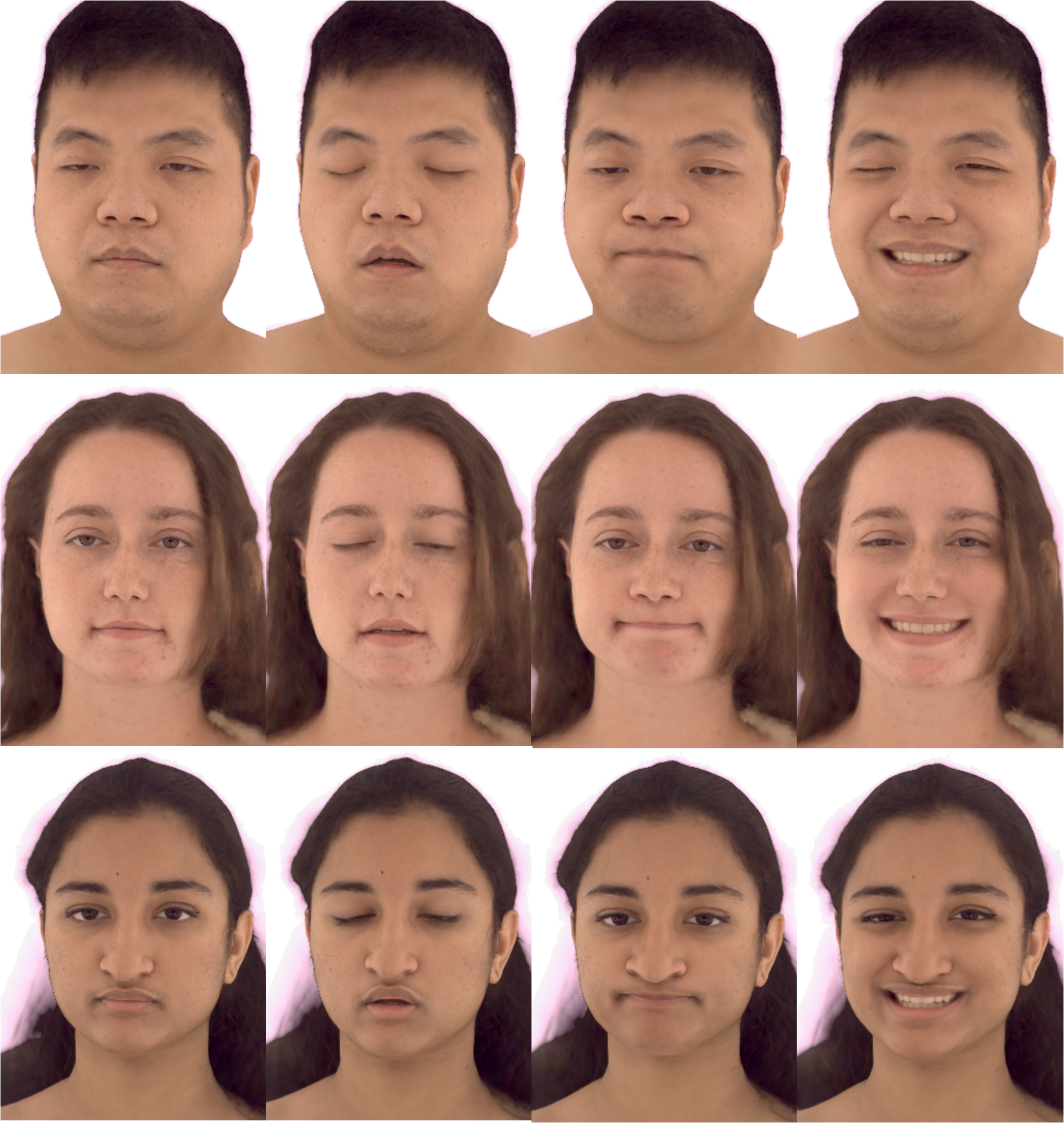}
\caption{\textbf{Conditioning Expressions for $\mathcal{S}$:} Four conditioning expressions $(\bs z_{k_1}, ..., \bs z_{k_4})$ for three different identities.}
\label{fig:cond_expr}
\end{figure}

The style transfer module, $\mathcal{S}$, has $\sim$25M trainable parameters and operates at an image resolution of $192\times 192$.  Both the input encoder and the conditioning encoder, as well as the decoder, follow the UNet architecture. We train a single style transfer network for all camera views by incorporating a learnable view embedding at each layer of the UNet.
Since the conditioning images are generated using the avatar model, $\mathcal{D}$, we also have access to their foreground masks and projected UV images of their guide mesh~\cite{Lombardi2021mvp}, which are also input to the conditioning encoder along with the rendered images.

\begin{figure}[t]
\centering
\includegraphics[width=0.9\textwidth]{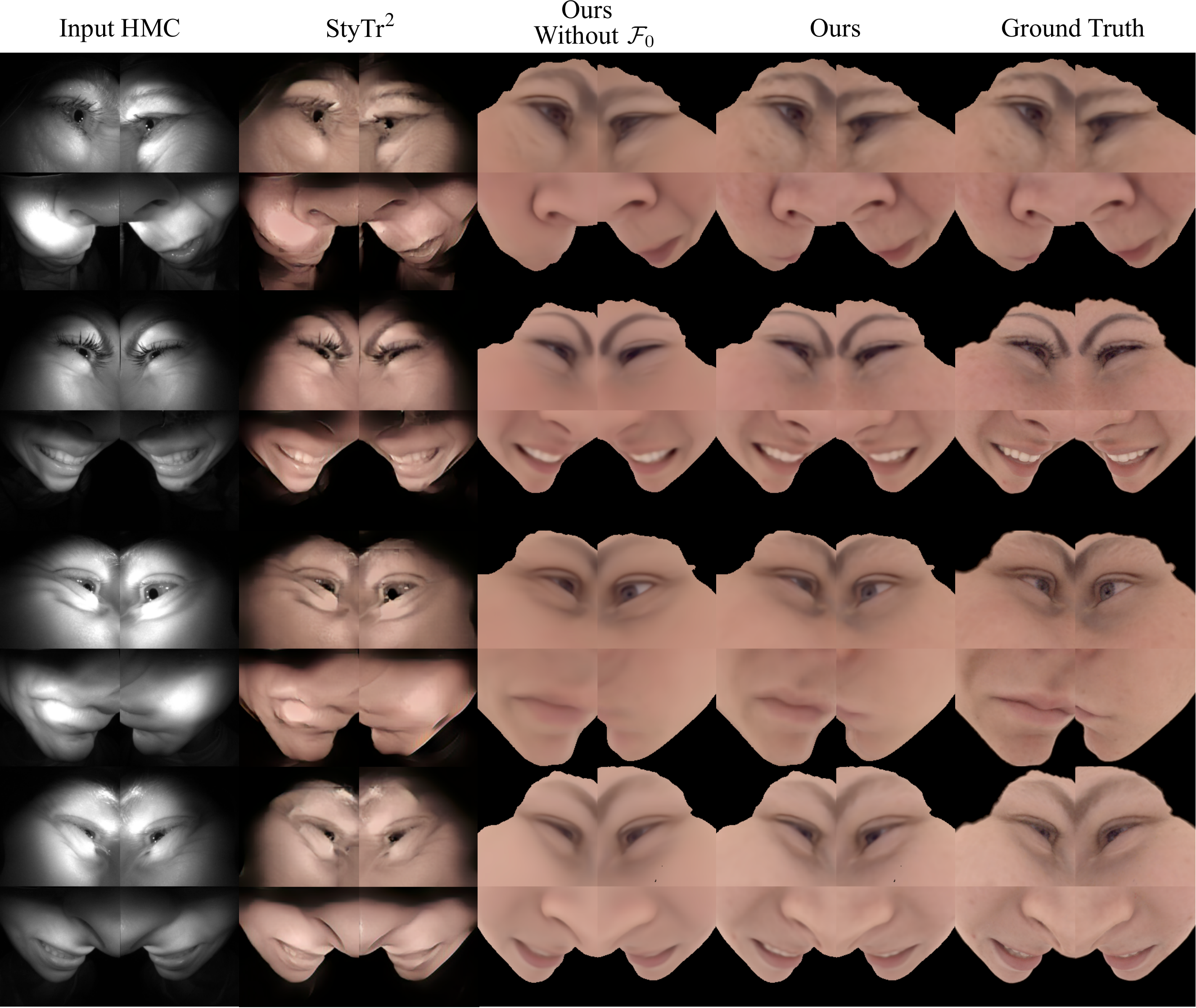}
\caption{\textbf{More Qualitative Results on Style Transfer:} We compare our results with a generic style transfer method as well as with our baseline method without the estimates by $\mathcal{F}_0$.}
\label{fig:more_qualitative_st}
\end{figure}

Fig.~\ref{fig:cond_expr} illustrates the four key conditioning expressions $(\bs z_{k_1}, ..., \bs z_{k_4})$ utilized in our experiments. These expressions were selected to cover extremes of the expression space, to compensate for information deficiency in style transfer conditioning while the estimate $\bs z_0$ is suboptimal.
Sliding Window Attention (SWA)~\cite{sasa_parmar2019stand} is based on the cross-attention layer of the transformer where each grid feature of the input branch cross-attends to a $5\times 5$ neighborhood around the aligned feature of the conditioning branch. SWA compensates for missregistration when the estimate $\bs v_0$ is suboptimal. We show more style transfer results on unseen test identities in Fig.~\ref{fig:more_qualitative_st}.

\section{HMC Details}
\begin{figure}[t]
\centering
\includegraphics[width=0.7\columnwidth]{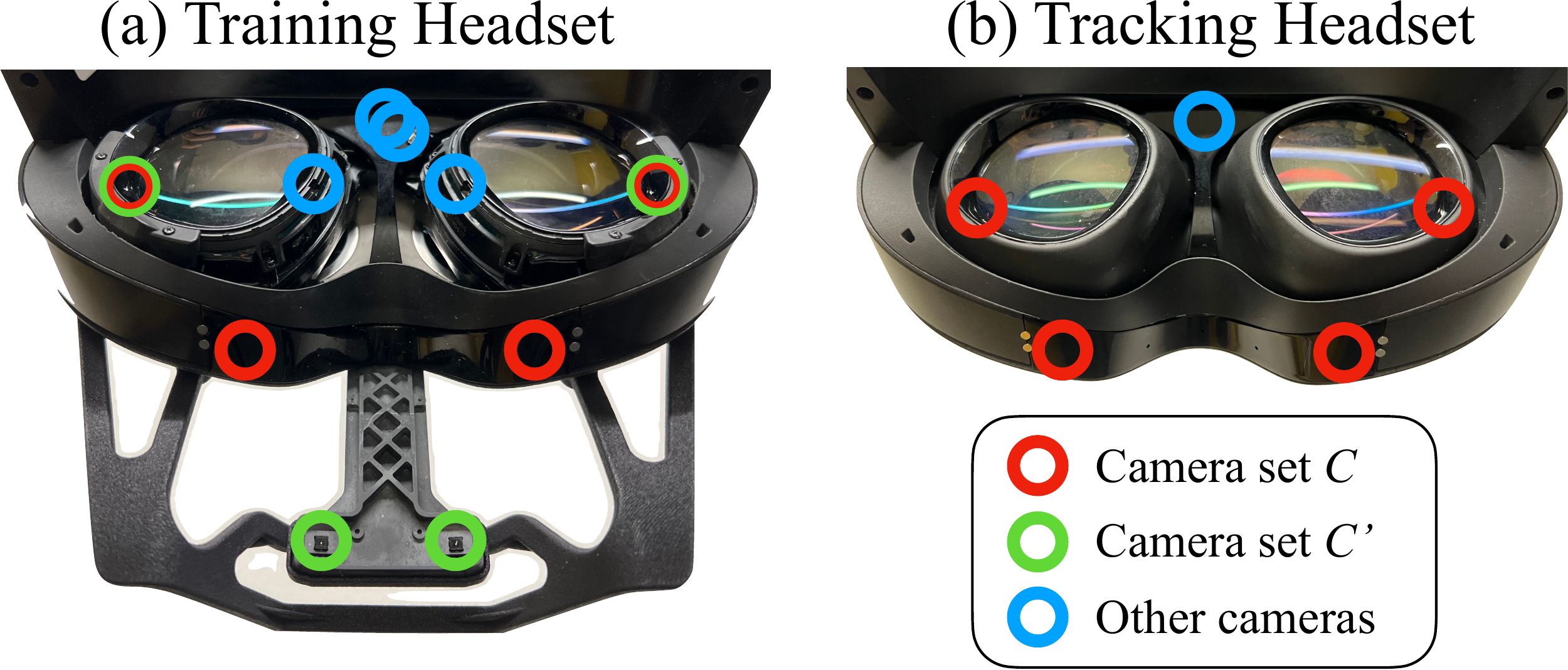}
\caption{\textbf{HMC details:} We use all cameras on training headset to establish ground truth in this work. Camera sets $C$ and $C'$ used in the main paper are annotated.}
\label{fig:hmc}
\end{figure}

In this work, we follow the concept of \textit{training headset} and \textit{tracking headset} proposed in~\cite{wei2019VR}, where the former has a superset of cameras of the latter (see Fig.~\ref{fig:hmc}). In this work, 
we use a more recent and advanced VR consumer headset QuestPro~\cite{questpro} as the tracking headset, and augment it with additional cameras on an extended structure as the training headset.
As shown in Fig.~\ref{fig:hmc} (a), there are 10 cameras on the training headset. We use all of them to establish ground truth with the method in~\cite{schwartz2020eyes}. 
Camera set $C$ on the tracking headset and the constructed camera set $C'$ used for comparison in the main paper are also annotated in the Fig.~\ref{fig:hmc}. Note we exclude the cyclopean camera on the tracking headset from the camera set $C$ due to limited observation and extreme illumination. We also focus on mouth area and did not compare against the other 2 eye cameras on the training headset. All cameras are synchronized and capture at 72 fps. 

\section{Training Details}

Our model is trained in phases, where $\mathcal{F}_0$ is first trained, followed by $\mathcal{S}$, which takes the pre-trained $\mathcal{F}_0$'s output as input. 
The error distribution of the estimates $(\bs z_0, \bs v_0)$ provided by $\mathcal{F}_0$ to $\mathcal{S}$ will vary between training and testing due to the generalization gap inherent in $\mathcal{F}_0$.
To address this discrepancy, we introduce random Gaussian noise to the estimates when training $\mathcal{S}$.
Similarly, we add random Gaussian noise the the prediction of $\mathcal{S}$ when training $\mathcal{F}$.
$\mathcal{F}$ is trained for $T=3$ refinement iterations. 
To stabilize training
the gradients of each iteration are not backpropagated to prior iterations;
we detach the predictions $(\bs z_{t+1}, \bs v_{t+1})$ before passing them as input to the next iteration.

Both $\mathcal{F}$ and $\mathcal{F}_0$ are trained for 200K steps with a minibatch size of 4 using the RAdam optimizer~\cite{radam_liu2019variance}. 
Weight decay is set to $10^{-4}$, and the initial learning rate is set to $3 \times 10^{-4}$. This learning rate is then gradually decayed to $3 \times 10^{-6}$ using a cosine scheduler. $\mathcal{S}$ is trained similarly except that the weight decay is set to $3 \times 10^{-4}$. The rotation component of viewpoint $\bs v$ is converted to a 6D-rotation representation~\cite{Zhou_2019_CVPR} before passing it to the network. Both loss weights $\lambda_{\text{hmc}}$ and $\lambda_{\text{front}}$ are set to 1.


\end{document}